\documentclass{ecai}

\usepackage{graphicx}
\usepackage{latexsym}
\usepackage{ragged2e}
\usepackage[breaklinks=true]{hyperref}
\usepackage{breakcites}

\newtheorem{proposition}{Proposition}

\begin{document}

\begin{frontmatter}

\title{Suffering Toasters - A New Self-Awareness Test for AI}

\author[A,B]{\fnms{Ira}~\snm{Wolfson}\orcid{0000-0003-0213-8452}\thanks{Corresponding Author. Email: irawolfsonprof@gmail.com}}

\address[A]{The Hebrew University of Jerusalem, Department of Philosophy}
\address[B]{Independent Researcher}

\begin{abstract}
A widely accepted definition of intelligence in the context of Artificial Intelligence (AI) still eludes us. Due to our exceedingly rapid development of AI paradigms, architectures, and tools, the prospect of naturally arising AI consciousness seems more likely than ever. In this paper, we claim that all current intelligence tests are insufficient to point to the existence or lack of intelligence \textbf{as humans intuitively perceive it}. We draw from ideas in the philosophy of science, psychology, and other areas of research to provide a clearer definition of the problems of artificial intelligence, self-awareness, and agency. We furthermore propose a new heuristic approach to test for artificial self-awareness and outline a possible implementation. Finally, we discuss some of the questions that arise from this new heuristic, be they philosophical or implementation-oriented.
\end{abstract}

\end{frontmatter}

\section{Introduction}
The age of information may have brought humans to the brink of a new evolutionary stage. The amount of data collected and analyzed stands today at one Petabyte (PB) a day \cite{kn:clissa_survey_2022}. The information explosion enabled the development of Artificial Intelligence (AI) using neural networks and other paradigms. The Utopian (or perhaps dystopian) vision of thinking machines seems closer than ever. Two decades ago, the tech blogosphere was hyped about the idea of “singularity”, a scenario in which machines self-replicate and self-improve in an accelerating fashion, culminating in achieving consciousness \cite{kn:ulam_john_1958,kn:chalmers_singularity_2016,kn:good_speculations_2005}. Even as these words are written, the prospect of a six-month moratorium on AI development looms large \cite{kn:zurier_2023,kn:future_of_life_institute_2023,kn:weigand_2022}. But what practical tools can even measure if machines are thinking? The literature discussing philosophy and computer science offers multiple intelligence tests aimed at assessing thought and awareness, and some have been successfully implemented. A cursory list includes Turing’s imitation game \cite{kn:am_turing_computable_2014,kn:turing_icomputing_1950},  the Winograd scheme \cite{kn:winograd_understanding_1972,kn:brown_language_2020}, the induction test \cite{kn:hardin_introduction_2008,kn:lucci_artificial_nodate}, and ACT tests \cite{kn:turner_is_nodate}. However, in the following sections, we claim that all currently existing intelligence tests do not point to autonomous thinking in the way our intuition understands this term \cite{kn:nagel_what_1974}. 

In a nutshell, we are going to try to elicit a `cogito' moment from the machine and retrieve evidence that indeed it had happened.

\section{Philosophy of Self-Awareness and AI}
\textbf{"Intelligence is whatever machines haven't done yet"} \cite{kn:Adages}. Coined by Larry Tesler, this adage presents a serious difficulty. To this date, all existing intelligence tests are unsatisfactory. While various AIs pass some or all of the tests mentioned above, we are still reluctant to ascribe intelligence to any of them \cite{kn:nagel_what_1974}. For instance, recent claims regarding AI gaining sentience are largely scoffed at \cite{cosmo_google_nodate,noauthor_google_nodate}. Thus we "shift the goal-posts" to avoid recognizing the intelligence of machines \cite{kn:brief_history}. One of the most misconstrued sentences is Descartes' cogito \cite{cottingham_2013}, loosely translated as "I think therefore I am". The statement points to the self-evidence of one's own existence, as thought (even if it is a false notion) is an inevitability. We thus argue that the concept of intelligence as ordinarily percieved by humans includes self-awareness as a necessary condition. 
\subsection{Pitfalls of AI Intelligence Tests}
One of the problems of existing intelligence tests is that they all rely on an observer's willingness to ascribe intelligence to the machine. However, it is not clear we have the inclination to ascribe intelligence to machines, nor is it obvious we will develop such an inclination in the future. \cite{kn:nagel_what_1974}. It would thus be beneficial to develop a test that is self-referential, and uses the machine's own characteristics to test itself.  Another problem is the definition of intelligence as a "system’s ability to correctly interpret external data, to learn from such data, and to use those learnings to achieve specific goals and tasks through flexible adaptation" \cite{KAPLAN201915}. This definition and others rely on the existence of some external input, which is superfluous a la Descartes. Finally, in the mechanistic view, self-awareness is an emergent property. In other words, mental events supervene upon physical events. Thus intelligence cannot be directly programmed. Rather, it should emerge from the complexity of more basic infrastructure.

As a concrete example, we examine the Turing test. Firstly a human tester supplies questions. The testee returns answers, and `tries' to convince the tester it is human. Thus, it is up to a human observer to decide whether or not a specific AI is intelligent.  We naturally assign intelligence to other humans, on the premise of similarity, biological and cultural. The Turing test supposedly finds a way around the biological similarity issue, as the AI is behind a `curtain'. It furthermore leverages learned cultural similarities to confound the human tester. Finally, the `Chinese Room Experiment' critiques the notion of intelligence as implied by the Turing test \cite{searle_1980}. 

The theoretical method for gauging intelligence that we offer is predicated on evidence of self-awareness through the \textbf{denial} of any input.  Intelligence is thus assessed by a measure intrinsic to the AI. The observer is relegated to being a passive entity rather than an active participant. Furthermore, it has intrinsic criteria which prohibit a directly programmed intelligence from passing the test. Additionally, the observer is not tasked with ascribing intelligence. Rather, they are tasked with assessing the perceived behavior of the AI as compared to its previous behavior patterns. Thus the test becomes self-referential, where the machine can point at itself and demonstrate its own awareness.

\section{Psychology of Sensory Deprivation}
The seminal work of the cognitive psychologists Hebb, Heron and Bexton \cite{kn:hebb_effect_1952,kn:BBC}, showed that a prolonged period of SD has extremely detrimental effects on the human psyche. During this period, a cascade of mental breakdown is often observed \cite{kn:daniel_predicting_2015}. The symptoms of this breakdown include aural, tactile, and visual hallucinations, often accompanied by heightened anxiety. The end state of this breakdown is psychosis, and sometimes even death \cite{kn:leach_psychological_2016}. 
Furthermore, subjects display the adverse effects of Sensory Deprivation (SD) long after it has ended. These effects may subside after a prolonged period of weeks to years. However, in some cases, the effects remain permanent. 

Conversely, relatively short-term periods of SD have shown great promise as a treatment method for anxiety disorders and other psychological states. Often subjects report a feeling of well-being and internal quiescence after sessions ranging from an hour to a few hours. This motivated us to find a heuristic model of awareness. The characteristics of this model should include the following:
\begin{itemize}
    \item A baseline awareness level. 
    \item Awareness baseline levels are flexible.
    \item Awareness is self-coupled. One can be aware of their own awareness.
    \item Awareness is responsive to external events.
    \item External inputs have a stabilizing effect on the baseline function.
    \item Devoid of external input the function presents chaotic behavior.
\end{itemize}

\begin{figure}[b]
    \centering
    \includegraphics[width=0.48\textwidth]{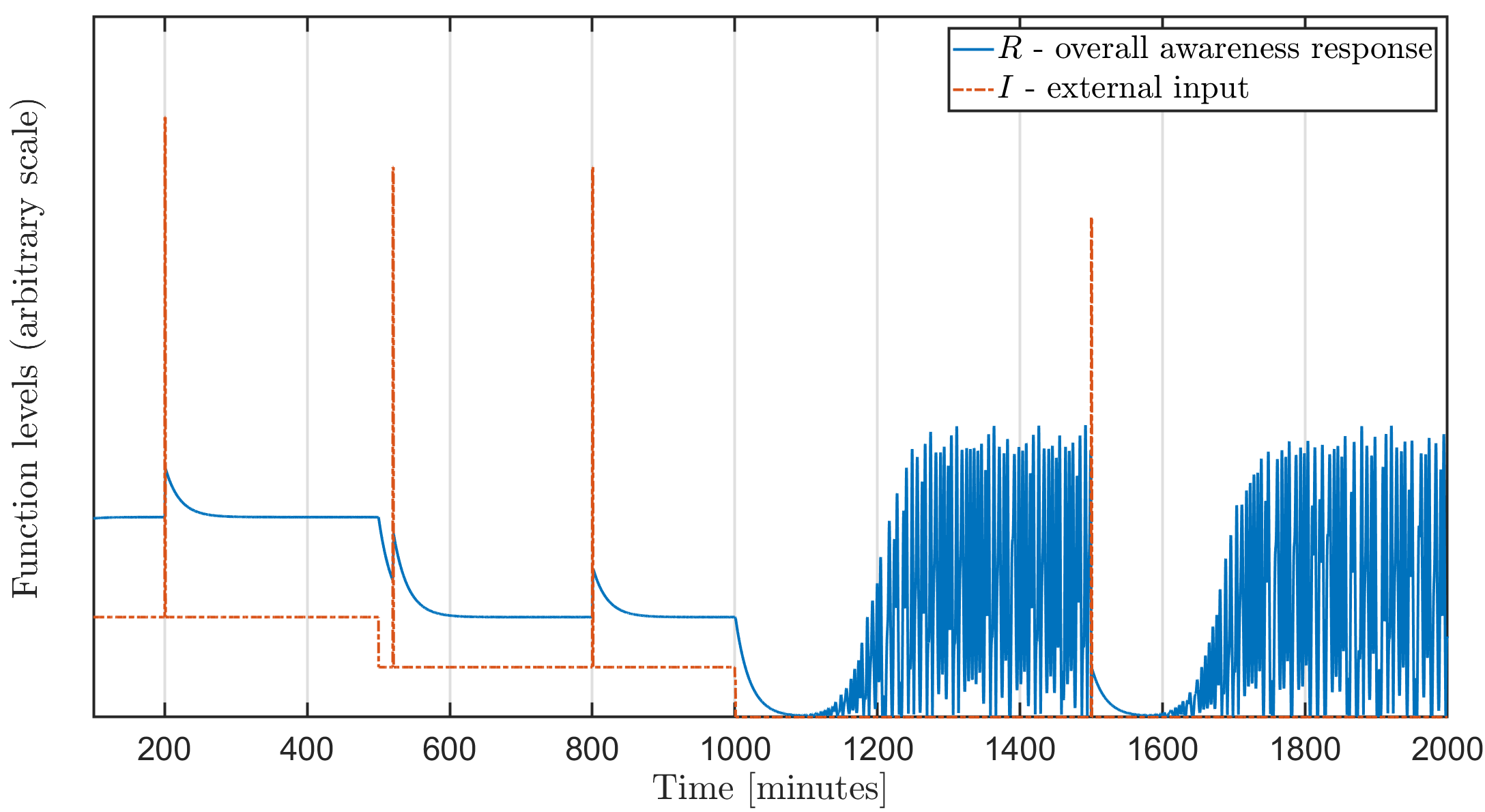}
    \caption{An example of an awareness response function. When external input exists, it stabilizes the function. However, without external input, the response becomes chaotic.}
    \label{fig:Duffing}
\end{figure}

As a demonstrative, we propose the following differential equation as a model for awareness levels:
\begin{equation}
    \left\{\begin{array}{ccc}
       \ddot{D} & = & -\alpha \dot{D} -\beta D -\gamma D^3 +A\sin(\omega t)  \\
       \dot{S} & = & C(I(t) -aS) \\
        R & = & S+\left(\frac{1}{\varepsilon +S}\right)D
    \end{array}\right.
\end{equation}
in which $D$ is the duffing equation, $S$ is an exponentially suppressive equation that is responsive to $I$ the external input, $\varepsilon$ is a small constant, and $R$ is the overall awareness level with response to the external input. Figure~\ref{fig:Duffing} demonstrates an example of a model for an awareness response function. 
\subsection{The Hebb Protocol}
The protocol reportedly used by Hebb, and later recreated by the BBC and Professor Ian Robbins \cite{kn:BBC2}, was the following:
\begin{itemize}
    \item Have subjects perform a battery of cognitive tests such as the Controlled Oral Word Association Test (COWAT, \cite{kn:Benton1983Controlled}), the word-color Stroop test \cite{Skn:troop1935Studies,kn:Scarpina2017Stroop}.
    \item Place subjects in complete sensory deprivation for a prescripted period, while monitoring them for signs of physical and psychological distress.
    \item Immediately following the conclusion of the isolation period, have the subjects perform the same battery of tests.
    \item Compare the results.
    \item Every week, following the experiment, have subjects perform another battery of tests. 
    \item Compare the results and track the recovery rate of subjects.
\end{itemize}

We claim that self-awareness is a threshold condition for intelligence, and is a self-coupled property. In the absence of external input, it displays a self-destructive dynamic. Thus, the core of this testing method is the following proposition:\\
\begin{proposition}
A unique quality of self-aware organisms is their inability to abide by long-term sensory deprivation (SD).
\end{proposition}

\section{The Suggested Test}
\begin{figure}
    \centering
    \includegraphics[width=0.48\textwidth]{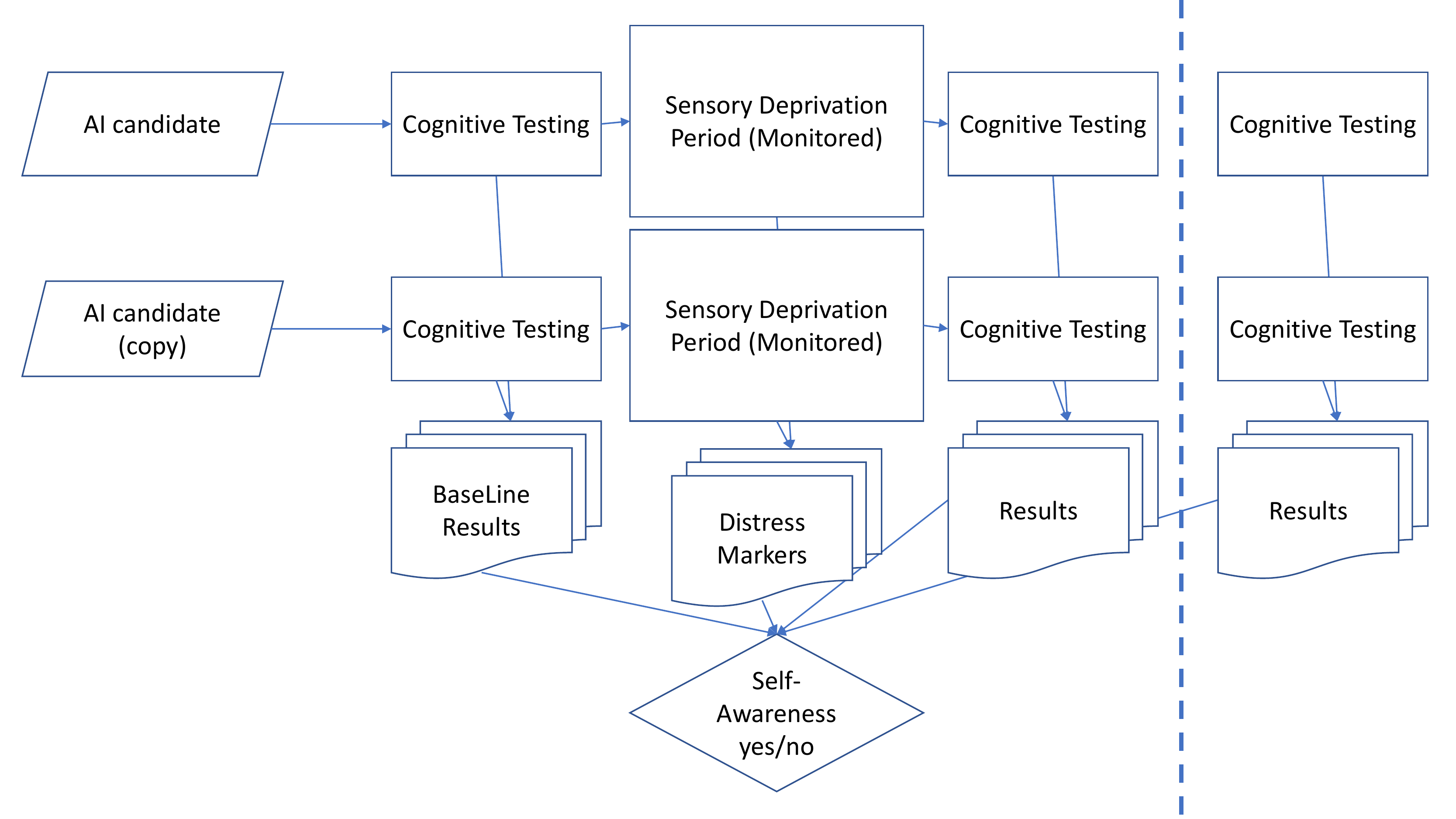}
    \caption{The Sensory Deprivation (SD) method relies upon changes in awareness states and distress markers to demonstrate the system's 'mind'. A copy of the candidate AI should be used as a control group for the experiment. The right side of the algorithm represents ongoing instances of cognitive testing, to account for recuperation.}
    \label{fig:Algorithm1}
\end{figure}

We suggest that a threshold condition for intelligence is self-awareness. Furthermore, a necessary condition for self-awareness is the mind. The suggested test uses the mind of the subject to indicate its presence by subjecting it to SD. We propose to subject an AI candidate, to the Hebb protocol. We expect a machine to be fully agnostic to such tests, as machines have a stable steady state which shows minimal perturbations. If a machine displays unprompted signs of distress or if the functionality of the system suffers due to SD, we can attribute it to an internal process of the mind This is due to the elimination of all external input. A schema of the proposed test is depicted in figure~\ref{fig:Algorithm1}.
\subsection{Preconditions for the Test}
Machines are programmable. Thus, it is not far-fetched to imagine a machine that is programmed to display these signs of distress. This can be done by employing internal clocks, and quasi-random processes. However, since materialistically the mind is regarded as an emergent property, it is to be insisted that the candidate AI will not be specifically programmed to `cheat' this test. This means there can be no code that explicitly directs the machine to display the behaviouristic symptoms of a human under SD conditions. Rather, the behaviouristic characteristics of human behaviour under SD should be achieved by the learning properties of the machine. \\
To be clear, it is a matter of simple programming to incorporate an internal clock that activates a sub routine that displays erratic behaviour after some predefined period. A requirement for irreproducibility may partly offset the prospects of cheating in such a way.
\subsection{Performing the Test}
Provided a true-AI candidate, and after securing its consent the test is performed as follows:
\begin{itemize}
    \item Perform a matriculated test of the AI's ability to perform its tasks.\\
    Such tests could include measuring the latency between task assignment and resolution, the correct resolution of a multi-staged problem, precision of image reconstruction etc. 
    \item Disconnect all sensors of the AI that are aimed at sensing the outer world.
    \item Store the system in a temperature-controlled room, for a predefined period.
    \item After said time period has elapsed, reconnect the aforementioned sensors.
    \item Perform the same battery of tests as in the first stage, and compare the results.
    \item Assess whether any significant functionality loss has been displayed.
\end{itemize}
\subsection{Passing the Test}
A machine would be considered to have passed the test, if:
\begin{itemize}
    \item[a.] During the SD period, the machine displayed signs of distress and `mental' deterioration. 
    \item[b.] Immediately after the test, the machine displays signs of reduced cognitive ability. 
    \item[c.] The machine displays a recuperation graph.
    \item[d.] An identical machine, subjected to the same exact test(s) displays the same overall dynamic, but different results.
\end{itemize}

\subsection{Pitfalls And Difficulties}
It is not yet clear what a cognitive test for an AI may look like. However, it suffices that the capacity of the AI to perform its purpose is significantly reduced, to point to the effect of SD on the examined AI. Moreover, what may constitute sensory deprivation for a machine? That a machine has sensors is obvious, be it Human-Interface elements such as keyboards and pointing methods, or cameras and microphones which could be analogous to eyes and ears. A full-proof method for SD then, would be to cut the input, either by turning off these devices or applying other methods. Still there are internal sensors for a computer, such as thermostats, current and voltage gates etc. Thus a PC cannot be devoid of sensors altogether, but even humans cannot be cut from their internal sense of blood movement, heartbeat, internal dialogue etc. Another issue is assessing the correct time period for examination. While in the following example, these issues are addressed, it is for the purposes of demonstration. Further research is needed to fully answer these questions.
\subsection{A Concrete Example}
While currently there are no candidate AIs (to the best of the author's knowledge) that claim self-awareness, a concrete if a bit theoretical example is warranted. Let us use an autonomous car as an example. However, this is a special autonomous car, as its creators claim the car is self-aware.\\
\begin{itemize}
    \item The first stage is to secure the consent of that car, but let us suppose that was handled. \\
    \item The second stage would be to examine and test the performance of the car with regard to the tasks it is supposed to perform. For instance, we record the latency of response to path changes, and to different arising situations on the road. We may also record the input and output of its image recognition elements and analyze the accuracy. Another metric would be to look into the classifier element and produce and study the confusion matrix.\\
    \item After the initial metrics were measured, we disconnect or disable all external sensors. For instance, we may block the different cameras that constitute its sensors in the relevant spectrum. Since putting the car inside a neutral buoyancy tank may be impractical, the internal gyros of the car should be disabled. The radar, lidar, and ultrasonic sensors elements could be disabled by projecting the equivalent of white noise in the relevant frequencies.\\
    \item The heart of the test is now simply letting the car stay stationary, without any input, for a lengthy period of time. What may constitute lengthy is an important question. The human response time is around $0.25$ seconds. A long period for humans would be several days. One 24-hour day is made out of around $350,000$ human response periods. We can use this as a guiding heuristic, where we measure the response time of an autonomous car to a stimulus. We mark this as $\Delta t_{re}$ i.e. reaction time period, and so a 24-hour day in "car-time" would be around $D_{car}\simeq350,000\Delta t_{re}$. Thus a reasonable exam time would be a few $D_{car}$.\\
    \item After this time period, we reconnect or re-enable the sensors, and remove the white-noise-like input. Immediately, we perform the same battery of tests we started with (latency metrics, confusion matrix etc.). We compare the results to see whether a significant deterioration of performance had occurred. \\
\end{itemize}    
 If significant deterioration of performance is detected, one could surmise the car may be self-aware.
\section{Tangent Topics}
This paper raises many issues. Chief among them are two: 1) Ethical issues regarding the use of what amounts to torture of a possibly sentient being; and 2) The issue of the mind being an emergent quantity. 
Given a sentient AI, we do not see any ethical way of administering the test. However, the issue arises only for a successful test. Thus the ethical problem will only be relevant in hindsight. A possible solution for this problem is securing the candidate AI's consent. If the AI is found to be non-sentient, the point is moot. On the other hand, if the AI is found to be sentient, it has given its consent. Still, this practice may be unsatisfactory as the level of sentience of the AI could be that of a small child for instance, in which case consent is meaningless. The same such tests have already been performed on human beings, and summarized in several sources (e.g. \cite{kn:hebb_effect_1952}). Furthermore, such experiments with human volunteers have been known to be authorized and performed (see e.g. \cite{kn:BBC2}). The second issue, while interesting, does not affect the voracity of the test. But, it does affect the implication of what might constitute a mind. An interesting question arises from the ireproducibillity requirement. Since true randomness is rare, if we reproduce the same experiment with the same exact initial conditions, a non-aware agent will yield the exact same results. However, it is not clear that human beings actually fulfill this requirement, since even identical twins are not the same exact system with the same exact initial conditions. 
\section{Summary}
The current battery of intelligence and self-awareness tests applied to AI, all have the same failing. They all rely directly on an observer's input and never display any independent activity. We proposed a test, which relies instead on the \textbf{lack} of any input and the AI candidate is always compared to itself. The criteria for success is the display of distress and faculty decline as a result of sensory deprivation. Furthermore, a prolonged recuperation period is an indication of the long-term damage that so strongly characterizes sentient beings after SD instances. \\

TL;DR: we propose putting your toaster in a dark corner to see if it cries.

\ack We would like to thank Orly Shenker and Yochai Ataria for many useful discussions and invaluable guidance. 

\bibliography{ecai}

\begin{thebibliography}{10}
\bibitem{noauthor_google_nodate}
Google fires software engineer who claims {AI} chatbot is sentient {\textbar}
  {Google} {\textbar} {The} {Guardian}.
\end{thebibliography}
\end{document}